\documentclass[10pt,twocolumn,letterpaper]{article}
\usepackage[accsupp]{axessibility}
\usepackage{iccv}
\usepackage{times}
\usepackage{epsfig}
\usepackage{graphicx}
\usepackage{amsmath}
\usepackage{amssymb}
\usepackage{algorithm}
\usepackage{algorithmic}
\usepackage{color}

\usepackage{amsmath,amssymb}
\usepackage{multirow}
\usepackage{booktabs}
\def\rvx{{\mathbf{x}}}

\def\rvz{{\mathbf{z}}}
\def\N{{\mathcal{N}}}

\def\Z{{\mathbb{Z}}}

\def\p{p_{\rm data}}
\newcommand\blfootnote[1]{%
  \begingroup
  \renewcommand\thefootnote{}\footnote{#1}%
  \addtocounter{footnote}{-1}%
  \endgroup
}

\usepackage[pagebackref=true,breaklinks=true,letterpaper=true,colorlinks,bookmarks=false]{hyperref}

\iccvfinalcopy 

\def\iccvPaperID{12058} 

\ificcvfinal\fi

\begin{document}

\title{Learning Hierarchical Features with Joint Latent Space Energy-Based Prior}

\author{Jiali Cui$^1$, Ying Nian Wu$^2$, Tian Han$^1$\thanks{: corresponding author}\\
$^1$Department of Computer Science, Stevens Institute of Technology\\
$^2$Department of Statistics, University of California, Los Angeles\\
{\tt\small \{jcui7,than6\}@stevens.edu, ywu@stat.ucla.edu}
}

\maketitle
\ificcvfinal\thispagestyle{empty}\fi
\begin{abstract}
This paper studies the fundamental problem of multi-layer generator models in learning hierarchical representations. The multi-layer generator model that consists of multiple layers of latent variables organized in a top-down architecture tends to learn multiple levels of data abstraction. However, such multi-layer latent variables are typically parameterized to be Gaussian, which can be less informative in capturing complex abstractions, resulting in limited success in hierarchical representation learning. On the other hand, the energy-based (EBM) prior is known to be expressive in capturing the data regularities, but it often lacks the hierarchical structure to capture different levels of hierarchical representations. In this paper, we propose a joint latent space EBM prior model with multi-layer latent variables for effective hierarchical representation learning. We develop a variational joint learning scheme that seamlessly integrates an inference model for efficient inference. Our experiments demonstrate that the proposed joint EBM prior is effective and expressive in capturing hierarchical representations and modelling data distribution. \blfootnote{Our project page is available at \url{https://jcui1224.github.io/hierarchical-representation-ebm-proj/}.} 
\end{abstract}
\section{Introduction}
In recent years, deep generative models have achieved remarkable success in generating high-quality images \cite{vahdat2020nvae,song2020score}, texts \cite{yu2017seqgan,guo2018long}, and videos \cite{Saito_2017_ICCV,Tulyakov_2018_CVPR}. However, learning generative models that incorporate hierarchical structures still remains a challenge. Such hierarchical generative models can play a critical role in enabling explainable artificial intelligence, thus representing an important area of ongoing research. 

To tackle this challenge, various methods of learning hierarchical representations have been explored \cite{sonderby2016ladder,zhao2017learning,maaloe2019biva,child2020very,sonderby2016ladder}. These methods learn generative models with multiple layers of latent variables organized in a top-down architecture, which have shown the capability of learning increasingly abstract representations (e.g., general structures, classes) at the top layers while capturing low-level data features (e.g., colors, background) at the bottom layers. In general, one could categorize these methods into two classes: (i) conditional hierarchy and (ii) architectural hierarchy.

\textit{The conditional hierarchies} \cite{sonderby2016ladder,gulrajani2016pixelvae,nijkamp2020learning,maaloe2019biva,child2020very} employ stacked generative models layered on top of one another and assume conditional Gaussian distributions at different layers, while \textit{the architectural hierarchies} \cite{zhao2017learning,li2020progressive} instead leverage network architecture to place high-level representations at the top layers of latent variables and low-level representations at the bottom layers. However, they typically assume conditional Gaussian distribution or isotropic Gaussian as the prior, which could have limited expressivity \cite{pang2020learning}. Learning a more expressive and informative prior model for multiple layers of latent variables is thus needed. 

The energy-based models (EBMs) \cite{lecun2006tutorial,xie2016theory,du2019implicit,nijkamp2019learning,pang2020learning,cui2023learning}, on the other hand, are shown to be expressive and proved to be powerful in capturing data regularities. Notably, \cite{pang2020learning} studies the EBM in the latent space, where the energy function is considered as a correction of the non-informative prior and tilts the non-informative prior to a more expressive prior distribution. The low dimensionality of the latent space makes EBM more effective in capturing regularities in the data. However, prior methods often rely on expensive MCMC methods for posterior sampling of the latent variables, and more importantly, the latent variables are not hierarchically organized, making it difficult to capture data variations at different levels.

In this paper, we propose to combine the strengths of latent space EBM and the multi-layer generator model for effective hierarchical representation learning. In particular, we build the EBM on the latent variables across different layers of the generator model, where the latent variables at different layers are concatenated and modelled jointly. The latent variables at different layers capture different levels of information from the data via the hierarchy of the generative model, and their inter-relations are further tightened up and better captured through EBM in joint latent space. 

Learning the EBM in the latent space can be challenging since it usually requires MCMC sampling for both the prior and posterior distributions of the latent variables. The prior sampling is efficient with the low dimensionality of the latent space and the lightweight network of the energy function, while the MCMC sampling of posterior distribution requires the backward propagation of generation network, which is usually heavy and thus renders inefficient inference. To ensure efficient inference, we introduce an inference model and develop a joint training scheme to efficiently and effectively learn the latent space EBM and multi-layer generator model. In particular, the inference model aims to approximate the exact posterior of the multi-layer latent variables while also serving to extract levels of data variations through the deep feed-forward inference network to facilitate hierarchical learning.

\noindent\textbf{Contributions:} 1) We propose the joint EBM prior model for hierarchical generator models with multiple layers of latent variables; 2) We develop a variational training scheme for efficient learning and inference; 3) We provide strong empirical results on hierarchical learning and image modeling, as well as various ablation studies to illustrate the effectiveness of the proposed model.

\section{Background}
In this section, we present the background of multi-layer generator models and the latent space EBM, which will serve as the foundation of the proposed model.
\subsection{Multi-layer generator models}
\label{section-bg-cond}
Let $\rvx \in \mathbb{R}^D$ represent the high-dimensional observed examples, and $\rvz \in \mathbb{R}^d$ denote the low-dimensional latent variables. The multi-layer generator model \cite{sonderby2016ladder,vahdat2020nvae,maaloe2019biva} consists of multiple layers of latent variables that are organized in a top-down hierarchical structure and modelled to be conditionally dependent on its upper layer, i.e., $p_{\theta}(\rvz) = \prod_{i=1}^{L-1}p_{\theta_i}(\rvz_i|\rvz_{i+1})p_0(\rvz_L)$, where $p_{\theta_i}(\rvz_i|\rvz_{i+1})\sim \mathcal{N}(\mu_{\theta_i}(\rvz_{i+1}), \sigma_{\theta_i}(\rvz_{i+1}))$ and $p_0(\rvz_L) \sim \mathcal{N}(0, I)$. 
Although such a modelling strategy has seen significant improvement in data density estimation, it can be less effective in learning hierarchical representations.

\begin{figure}[h]
\centering 
\includegraphics[width=0.28\columnwidth]{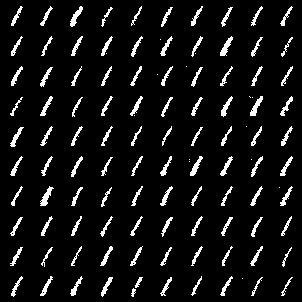}
\includegraphics[width=0.28\columnwidth]{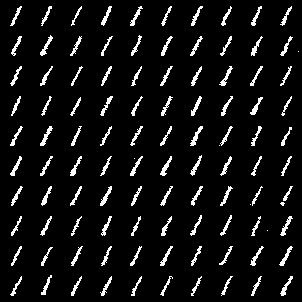} 
\includegraphics[width=0.28\columnwidth]{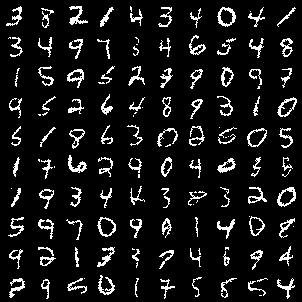} 
\caption{Visualization of hierarchical representations on BIVA. \textbf{Left:} sampling $\epsilon_1, \epsilon_2$ for bottom layers. \textbf{Middle:} sampling $\epsilon_3, \epsilon_4$ for middle layers. \textbf{Right:} sampling $\epsilon_5, \epsilon_6$ for top layers.} 
\label{Fig.biva-mnist}
\end{figure}

To demonstrate, we train BIVA\footnote{https://github.com/vlievin/biva-pytorch} \cite{maaloe2019biva} on MNIST with 6 layers of latent variables and conduct hierarchical sampling by generating latent variables via the repamaramization trick, i.e., $\rvz_i =\mu_{\theta_i}(\rvz_{i+1})+\sigma_{\theta_i}(\rvz_{i+1})\cdot\epsilon_i$, where $\epsilon_i$ is the added Gaussian noise. Specifically, the latent variables are generated by randomly sampling $\epsilon_i$ for the $i$-th layer, while fixing $\epsilon_{j\ne i}$ for other layers, and the corresponding image synthesis should deliver the variation of representation learned by $\rvz_i$. We show the results in Fig. \ref{Fig.biva-mnist}, where the major semantic variation is only captured by the top layer of the latent variables, suggesting an ineffective hierarchical representation learned (compare with our model shown in Fig. \ref{Fig.mnist}).

\vspace{-1em}

\subsection{Latent space EBM}
\label{section-bg-lebm}
\begin{figure}[h]
\centering 
\includegraphics[width=0.45\columnwidth]{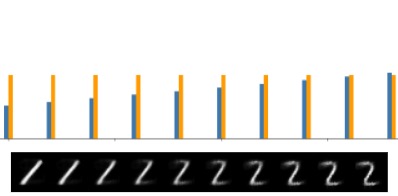}
\includegraphics[width=0.45\columnwidth]{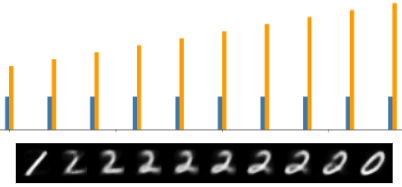}
\caption{Visualization for latent space EBM \cite{pang2020learning} by changing each unit of 2-dimensional $\rvz$. \textbf{Left}: changing the first unit. \textbf{Right}: changing the second unit. \textbf{Top:} the value of each unit, where the orange color indicates the first unit and the blue color indicates the second unit. } 
\label{Fig.lebm-mnist}
\end{figure}

The energy-based models (EBMs) \cite{du2019implicit,du2020improved,nijkamp2019learning,gao2020learning,xiao2020vaebm} present a flexible and appealing way for data modelling and are known for being expressive and powerful in capturing data regularities. Albeit their expressivity, learning the EBM on high-dimensional $\rvx$-space can be challenging since the energy function needs to support the entire data space. As such, another line of works \cite{pang2020learning, aneja2021contrastive,xiao2022adaptive} seek to build EBM on \textit{low-dimensional} $\rvz$-space with the probability density defined as 
\begin{eqnarray}
\label{lebm}
     p_{\theta}(\rvz) = \frac{1}{\Z(\theta)}\exp{[f_{\theta}(\rvz)]}p_0(\rvz)
\end{eqnarray}
However, with a single-layer non-hierarchical $\rvz$-space, it is typically \textit{infeasible} to capture the data abstraction of different levels. To see this, we pick up the digit classes `0', `1', and `2' of the MNIST dataset, on which we train LEBM \cite{pang2020learning} with the latent dimension set to be 2. We show in Fig. \ref{Fig.lebm-mnist} that changing the value of each latent unit could lead to a \textit{mixed} variations in digit shape and class (compare to our proposed method in Fig. \ref{Fig.ours-mnist}). 
\vspace{-1em}

\section{Methodology}
In this paper, we study a joint latent space EBM prior where latent variables are partitioned into multiple groups but modelled jointly for hierarchical representation learning. To ensure efficient inference, we introduce an inference model to approximate the generator posterior and develop a variational joint training scheme in which each model can be trained efficiently and effectively.
\subsection{Latent variable generative model}
A latent variable generative  model (i.e., generator model) can be specified using joint distribution:
\begin{equation}\label{our-joint}
p_{\beta, \alpha}(\rvx, \rvz) = p_{\beta}(\rvx|\rvz)p_\alpha(\rvz)
\end{equation}
which consists of the generation model $p_\beta(\rvx|\rvz)$ and the prior model $p_\alpha(\rvz)$.

\noindent\textbf{Joint latent space EBM:} The prior model $p_\alpha(\rvz)$ is usually assumed to be single-layer, which can be infeasible for learning hierarchical representations (see Sec. \ref{section-bg-lebm}). In this work, we propose a joint latent space EBM prior, where latent variables are separated into multiple groups, i.e., $\rvz = [\rvz_1, \rvz_2, \dots, \rvz_L]$, and our EBM prior model is defined as
\begin{align}\label{our-ebm}
p_{\alpha}(\rvz)=& \frac{1}{\Z(\alpha)}\exp[f_{\alpha}([\rvz_1, \dots, \rvz_{L}])] p_0([\rvz_1, \dots, \rvz_{L}])
\end{align}
where $[.]$ refers to the concatenation. $f_{\alpha}(.)$ is the negative energy function that can be parameterized by a small multi-layer perceptron with the parameters $\alpha$, $p_0([\dots])$ is a reference distribution assumed to be a unit Gaussian, and $\Z(\alpha)$ is the normalizing constant or partition function. Such a prior model can be considered as an energy-based refinement of the non-informative reference distribution, thus the relationship between different groups of latent codes can be well captured via the energy function $f_\alpha([\rvz_1, \dots, \rvz_{L}])])$. 

For notation simplicity, we denote $\rvz = [\rvz_1, \dots, \rvz_L]$ in subsequent discussions.  

\noindent \textbf{Generation Model.} The generation model is defined as
\begin{equation}
    p_{\beta}(\rvx|\rvz) \sim \mathcal N(g_{\beta}(\rvz), \sigma^2 I_D)\label{our-gen-p}
\end{equation}
which models the observed data using top-down generator network $g_\beta(\rvz)$ with additive Gaussian noise, i.e., $\rvx = g_\beta(\rvz) +  \epsilon, \epsilon \sim \mathcal N(0, \sigma^2 I_D)$. To facilitate the hierarchical representation learning with multi-layer latent variables, we consider multi-layer hierarchical generator network $g_{\beta}$ ($ = \{g_{1}, g_{2}, \dots, g_{L}\}$) that is designed to explain the observation $\rvx$ by integrating data representation from the above layers, i.e.,
\begin{align}\label{our-gen}
    &h_{L} = g_{L}(\rvz_{L})\nonumber \\ 
    &h_i=g_{i}([\rvz_i, h_{i+1}]),\, \, \, i=1, 2, \dots, L-1\\
    &\rvx \sim \mathcal N(h_1, \sigma^2 I_D)\nonumber 
\end{align}
in which $\rvz_L$ is at the top layer, and $g_i$ is a shallow network that decodes latent code $\rvz_i$ while integrating features from the upper layer. 

\noindent\textbf{Joint EBM prior vs. independent Gaussian prior:} With such a multi-layer hierarchical generation model, it is also tempting to consider learning with the unit Gaussian prior, i.e., $p(\rvz)\sim\N(0, I_d)$. Such a prior choice is adopted in \cite{zhao2017learning, li2020progressive} and has seen some success in learning hierarchical features. However, the independent Gaussian prior does not account for the relations between levels of representations and can be less expressive in capturing complex data features. The proposed EBM prior is expressive and can couple the latent variables across different layers within a joint modelling framework. \\

\begin{figure}[t]
\centering 
\includegraphics[width=0.99\columnwidth]{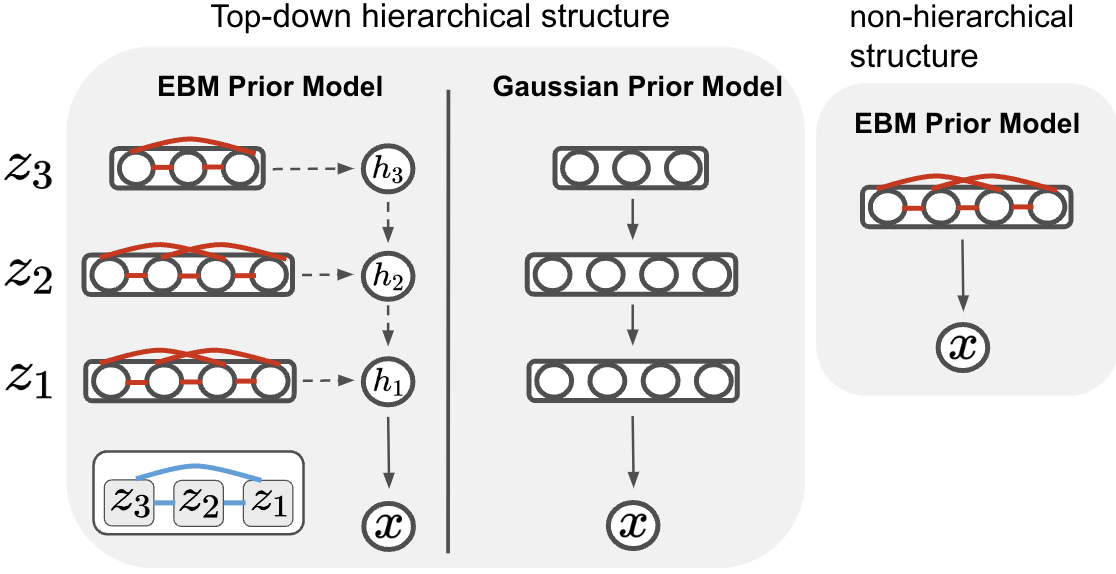} 
\caption{The illustration of the proposed joint EBM prior model (Left). \textbf{Red lines} indicates the modelling of \textit{intra-layer} relation, and \textbf{blue lines} indicate \textit{inter-layer} relation. Compared to the Gaussian prior model and single-layer LEBM, our prior model is expressive and effective in learning hierarchical representations. }
\label{Fig.model}
\end{figure}

Our model is illustrated in the left panel of Fig. \ref{Fig.model}.

\subsection{Variational learning scheme}
Suppose we have observed examples ($\rvx_i, i$ = $1,\dots, n$) that come from (true) data distribution $\p(\rvx)$ and denote $\theta=(\beta, \alpha)$ that collects all parameters from the generator model. Learning the generator model can be done by maximizing the log-likelihood of all observed examples as 
\begin{eqnarray}\label{log-likelihood}
\sum_{i=1}^{n} \log p_{\theta}(\rvx_i)= \sum_{i=1}^{n} \log \int p_\alpha(\rvz)p_\beta(\rvx_i|\rvz)d\rvz
\end{eqnarray}
When $n$ becomes sufficiently large, maximizing the above log-likelihood is equivalent to minimizing the Kullback-Leibler (KL) divergence between the model distribution and data distribution, i.e., $\min_\theta D_{\mathrm{KL}}(\p(\rvx)||p_{\theta}(\rvx))$. 

However, directly maximizing Eq. \ref{log-likelihood} can be challenging, as the inference of generator posterior (by chain rule, $p_\theta(\rvz|\rvx) = p_\theta(\rvx, \rvz) / p_\theta(\rvx)$) is typically intractable. To tackle the challenge, we employ an inference model $q_\phi(\rvz|\rvx)$ with a separate set of parameters $\phi$ to approximate the generator posterior. 

\noindent\textbf{Inference model:} The inference model maps from data space to latent space and is typically assumed to be Gaussian distributed, 
\begin{equation}
    q_\phi(\rvz|\rvx) \sim \N(\mu_\phi(\rvx), V_\phi(\rvx))\label{our-inf-q}
\end{equation}
where $\mu_\phi(\rvx)$ and $ V_\phi(\rvx)$ are the d-dimensional mean vector and diagonal covariance matrix. To match with the hierarchical generator model and facilitate the hierarchical representation learning, we consider the multi-layer inference model defined as
\begin{align}\label{inf_structure}
    &r_1 =h_1(\rvx) \nonumber \\ 
    &r_{i}=h_i(r_{i-1})\, \,  \, i=2, 3, \dots, L\\
    &\rvz_i \sim \mathcal N(\mu_i(r_i), V_i(r_i))\nonumber
\end{align}
where $\rvz_{L}$ is the inferred latent code at the top layer and $\rvz_1$ is the inferred bottom layer latent code. Each latent code $\rvz_i$ can be inferred and sampled through {reparametrization trick} \cite{kingma2013auto} using $\rvz_i = \mu_i(r_i) + V_i(r_i)^{1/2}\epsilon$ where $\epsilon \sim \mathcal N(0, I_{d_i})$, with $d_i$ being the dimensionality of $z_i$.

\noindent\textbf{Variational learning:} We show that the proposed generator model and the inference model can be naturally unified and jointly trained within a variational learning scheme. 

Specifically, we can define two joint densities, one for generator model density, i.e., $p_\theta(\rvx, \rvz)=p_\alpha(\rvz)p_\beta(\rvx|\rvz)$, and one for data density, i.e., $q_\phi(\rvx, \rvz) = \p(\rvx)q_\phi(\rvz|\rvx)$. We propose joint learning through KL minimization, and we denote the objective to be $L(\theta, \phi)$, i.e., 
\begin{eqnarray}\label{eq:kl_joint}
\min_\theta \min_\phi L(\theta, \phi) = \min_\theta \min_\phi D_{\mathrm{KL}}(q_{\phi}(\rvx, \rvz)||p_{\theta}(\rvx,\rvz))
\end{eqnarray}
where the generator model density is learned to match the true joint data density and capture extracted hierarchical features. 

\noindent\textbf{Learning $p_\alpha(\rvz)$:} With this joint KL minimization, the prior model can be learned with the gradient $-\nabla_\alpha L(\theta, \phi)$
\begin{align}
   &\mathbb{E}_{q_{\phi}(\rvz_1, \rvz_2, \dots, \rvz_L|\rvx)}[\nabla_{\alpha}f_{\alpha}([\rvz_1, \rvz_2, \dots, \rvz_L])] -\nonumber\\ 
   &\mathbb{E}_{p_{\alpha}(\rvz_1, \rvz_2, \dots, \rvz_L)}[\nabla_{\alpha}f_{\alpha}([\rvz_1, \rvz_2, \dots, \rvz_L])] \label{ebm_grad_new}
\end{align}
\noindent\textbf{Learning $p_\beta(\rvx|\rvz)$:} The generation model can be learned with the gradient $-\nabla_\beta L(\theta, \phi)$ computed as
\begin{equation}
  \mathbb{E}_{q_{\phi}(\rvz_1, \rvz_2, \dots, \rvz_L|\rvx)}[\nabla_{\beta}\log p_{\beta}(\rvx|\rvz_1, \rvz_2, \dots, \rvz_L)] \label{gen_grad_new}
\end{equation}
\noindent\textbf{Learning $q_\phi(\rvz|\rvx)$:} The inference model can be learned by computing the gradient $-\nabla_\phi L(\theta, \phi)$ as
\begin{align}\label{inf_grad_new}
   &\nabla_{\phi}\mathbb{E}_{q_{\phi}(\rvz_1, \rvz_2, \dots, \rvz_L|\rvx)}[\log p_{\beta}(\rvx|\rvz_1, \rvz_2, \dots, \rvz_L)] -\\
   &\nabla_{\phi} D_{\mathrm{KL}}(q_{\phi}(\rvz_1, \rvz_2, \dots, \rvz_L|\rvx)||p_{\alpha}(\rvz_1, \rvz_2, \dots, \rvz_L))\nonumber
\end{align}

We provide further theoretical understanding in Sec. \ref{sec-theorectical}.

\subsection{Prior sampling} 
Training the proposed joint EBM prior model requires sampling $\rvz_1, \rvz_2, \dots, \rvz_L$ from the proposed EBM prior (see Eq. \ref{ebm_grad_new}). To do so, we can use Langevin dynamic \cite{neal2011mcmc}.

For arbitrary target distribution $p(\rvz)$, the Langevin dynamic iterates:
\begin{equation}
\rvz_{t+1} = \rvz_{t} + \frac{s^2}{2}\nabla_\rvz \log p(\rvz_{t}) + s\epsilon_{t} \label{langevin}
\end{equation}
where $t$ indexes the time step of Langevin dynamics. $s$ is the step size and $\epsilon_t \sim \mathcal N(0, I_d)$. We run the noise-initialized Langevin dynamics for $K$ steps. Noted that as $s\rightarrow 0$, and $K \rightarrow \infty$, the distribution of $\rvz_t$ will converge to the target $p(\rvz)$ regardless of the initial distribution of $\rvz_0$ \cite{neal2011mcmc}.

Specifically, for prior sampling, we use the prior distribution $p_\alpha(\rvz)$ as our target distribution, and the gradient $\nabla_\rvz \log p_\alpha(\rvz)$ is computed as:
\begin{eqnarray}
\nabla_\rvz f_{\alpha}([\rvz_1, \dots, \rvz_{L}]) + \nabla_\rvz \log p_0([\rvz_1, \dots, \rvz_{L}])
\label{ebm_langevin_grad}
\end{eqnarray}
The Langevin dynamics for prior sampling can be efficient due to the low-dimensional latent space and the lightweight network structure for the energy function. \\

The overall procedure is summarized in Algorithm \ref{alg}.
\begin{algorithm}[!]
   \caption{Learning latent EBM prior for hierarchical generator model}
   \label{alg}
\begin{algorithmic}
   \STATE {\bfseries Input:} Training iterations $T$, observed training examples $\{x_i\}_{i=1}^n$, batch size $m$, network parameters $\alpha, \beta, \phi$, learning rate $\eta_{\alpha}, \eta_{\beta}, \eta_{\phi}$, Langevin steps $K$, Langevin step size $s$, 
   \STATE Let $t = 0$ and $\theta=(\alpha, \beta)$;
   \REPEAT
   \STATE \textbf{Prior sampling} for $\{z^-_i\}_{i=1}^{m}$ using Eq. \ref{langevin} and Eq. \ref{ebm_langevin_grad} with $K$ and $s$.
   
   \STATE \textbf{Inference sampling} for $\{z^+_i\}_{i=1}^{m}$ using Eq. \ref{inf_structure} with $\{x_i\}_{i=1}^m$ and reparametrization. 
   
   \STATE \textbf{Learn joint EBM prior:} Given $\{z^-_i,z^+_i\}_{i=1}^{m}$, update $\alpha=\alpha - \eta_{\alpha}\nabla_{\alpha}L(\theta, \phi)$ using Eq. \ref{ebm_grad_new} with $\eta_{\alpha}$
   
   \STATE \textbf{Learn generation model}:Given $\{x_i,z^+_i\}_{i=1}^{m}$ update $\beta=\beta - \eta_{\beta}\nabla_{\beta}L(\theta, \phi)$ using Eq. \ref{gen_grad_new} with $\eta_{\beta}$
   
   \STATE \textbf{Learn inference model}: Given $\{x_i,z^+_i\}_{i=1}^{m}$ update $\phi=\phi - \eta_{\phi}\nabla_{\phi}L(\theta, \phi)$ using Eq. \ref{inf_grad_new} with  $\eta_{\phi}$
   \STATE Let $t = t + 1$;
   \UNTIL{$t = T$}
\end{algorithmic}
\end{algorithm}

\subsection{Theorectical understanding}\label{sec-theorectical}
\noindent\textbf{Divergence perturbation and ELBO.}
The KL joint minimization (Eq. \ref{eq:kl_joint}) can be viewed as a surrogate of the MLE objective with the KL perturbation term,
\begin{eqnarray*}
&&D_{\mathrm{KL}}(q_{\phi}(\rvx, \rvz)||p_{\theta}(\rvx,\rvz)) \\
&=& D_{\mathrm{KL}}(p_{\rm data}(\rvx)||p_\theta(\rvx)) + D_{\mathrm{KL}}(q_\phi(\rvz|\rvx)||p_\theta(\rvz|\rvx))
\end{eqnarray*}
where $D_{\mathrm{KL}}(p_{\rm data}(\rvx)||p_\theta(\rvx))$ is the MLE loss function, and the perturbation term $D_{\mathrm{KL}}(q_\phi(\rvz|\rvx)||p_\theta(\rvz|\rvx))$ measures the KL-divergence between inference distribution and generator posterior. The inference $q_\phi(\rvz|\rvx)$ is learned to match the posterior distribution of the generator without expensive posterior sampling. In fact, such KL minimization in the joint space is equivalent to evidence lower bound (ELBO). To see this, noted that $D_{\mathrm{KL}}(p_{\rm data}(\rvx)||p_\theta(\rvx))= -H(p_{\rm data}(\rvx))-\mathbb E_{p_{\rm data}}[\log p_\theta(\rvx)]$ where $H(p_{\rm data}(\rvx))$ is the entropy of the empirical data distribution and can be treated as constant $C \equiv - H(p_{\rm data}(\rvx))$ w.r.t model parameters $\theta$. Then $L(\theta, \phi)$ is computed as
\begin{eqnarray*}
&&-\mathbb E_{p_{\rm data}}[\log p_\theta(\rvx)] + D_{\mathrm{KL}}(q_\phi(\rvz|\rvx) \| p_{\theta}(\rvz|\rvx)) + C \\
&=& \mathbb E_{p_{\rm data}} \left[ \mathbb E_{q_\phi(\rvz|\rvx)}\left( \log \frac{q_\phi(\rvz|\rvx)}{p_\theta(\rvz|\rvx)}\right) 
-  \log p_\theta(\rvx) \right] +C \\
&=& \mathbb E_{p_{\rm data}}[ - {\rm ELBO}(\rvx; \theta, \phi)] + C
\end{eqnarray*}
where 
\begin{eqnarray*}
{\rm ELBO}(\rvx; \theta, \phi) &=& \log p_\theta(\rvx) - \mathbb E_{q_\phi(\rvz|\rvx)}\left[ \log \frac{q_\phi(\rvz|\rvx)}{p_\theta(\rvz|\rvx)}\right]\\
&=& \mathbb E_{q_\phi(\rvz|\rvx)}\left[ \log \frac{p_\theta(\rvx,\rvz)}{q_\phi(\rvz|\rvx)} \right]
\end{eqnarray*}
and is referred to as evidence lower bound (ELBO) in the literature \cite{kingma2013auto}. Noted that the prior model (i.e., $p_\theta(\rvx, \rvz)=p_\beta(\rvx|\rvz)p_\alpha(\rvz)$) is now parameterized to be our joint EBM prior. Minimizing Eq. \ref{eq:kl_joint} is equivalent to maximizing the evidence lower bound of the log-likelihood. 

\noindent\textbf{Short run sampling and divergence perturbation.}
The learning of the energy-based model in Eq. \ref{ebm_grad_new} requires MCMC sampling from $p_\alpha(\rvz)$. We adopt the short-run Langevin dynamics (Eq. \ref{langevin}) with fixed $K$ steps to sample from prior $p_\alpha(\rvz)$ for efficient computation. Such prior sampling amounts to approximates the objective $L(\theta, \phi)$ in Eq. \ref{eq:kl_joint} with yet another KL perturbation term, i.e., 
\begin{eqnarray*}
\tilde{L}(\theta, \phi) = L(\theta, \phi) - D_{\mathrm{KL}}(\tilde{p}_{\alpha^{(t)}}(\rvz)|p_\alpha(\rvz))
\end{eqnarray*}
where $\tilde{p}_{\alpha^{(t)}}(\rvz)$ refers to the distribution of latent codes after $K$ steps of Langevin dynamics (Eq. \ref{langevin}) in the $t$-th iteration of the learning. The corresponding learning gradient for the prior model is thus
\begin{equation*}
   - \nabla_{\alpha}\tilde{L}(\theta, \phi) = \mathbb{E}_{q_{\phi}(\rvz|\rvx)}[\nabla_{\alpha}f_{\alpha}(\rvz)] - \mathbb{E}_{\tilde{p}_{\alpha^{(t)}}(\rvz)}[\nabla_{\alpha}f_{\alpha}(\rvz)] \label{ebm_grad_new_approx}
\end{equation*}
When $K$ is sufficiently large, the perturbation term $D_{\mathrm{KL}}(\tilde{p}_{\alpha^{(t)}}(\rvz)|p_\alpha(\rvz)) \rightarrow 0$. The update rule for the energy-based prior model (also Eq. \ref{ebm_grad_new}) can be interpreted as self-adversarial learning. The prior model $p_\alpha(\rvz)$ serves as both generator and discriminator if we compare it to GAN (generative adversarial networks) \cite{goodfellow2014generative}. In contrast to GAN, our learning follows MLE with divergence perturbations, which in general does not suffer from issues such as mode collapsing and instability, as it does not involve competition between two separate models. The energy-based model can thus be seen as its own adversary or its own critic. 
\section{Related Work}
\noindent \textbf{Generator model.} The generator model has a top-down network that maps the low-dimensional latent code to the observed data space. VAEs \cite{kingma2013auto,rezende2014stochastic,tolstikhin2017wasserstein,takida2022sq} proposes variational learning by introducing an approximation of the true intractable posterior, which allows a tractable bound on log-likelihood to be maximized. Another line of work \cite{han2017alternating,nijkamp2020learning} trains the generator model without inference model by using Langevin dynamics for generator posterior sampling. We follow the former approach and use the inference model for efficient training. 

\noindent\textbf{Generator model with informative prior.} The majority of the existing generator models assume that the latent code follows a simple and known prior distribution, such as isotropic Gaussian distribution. Such assumption may render an ineffective generator as observed in \cite{tomczak2018vae,dai2019diagnosing}. Our work utilizes the informative and expressive energy-based prior in the joint space of latent variables and is related to the line of previous research on introducing flexible prior distributions. For example, \cite{tomczak2018vae} parameterizes the prior based on the posterior inference model. \cite{dai2019diagnosing,ghosh2019variational} adopt a two-stage approach which first trains a latent variable model with simple prior and then trains a separate prior model (e.g., VAE or Gaussian mixture model) to match the aggregated posterior distribution. 

\noindent\textbf{Generator model with hierarchical structures.} The generator models that consist of multi-layers of latent variables are recently attracting attention for hierarchical learning. \cite{NIPS2016_6ae07dcb,bachman2016architecture,maaloe2019biva,child2020very,vahdat2020nvae} stack generative models on top of each other and assume conditional Gaussian distributions at different layers. However, as shown in Sec. \ref{section-bg-cond}, the bottom layer could act as a single generative model and absorb most features, leading to ineffective representation learning. \cite{zhao2017learning,li2020progressive} propose to learn hierarchical features with unit Gaussian prior by adapting the generator into multiple layers. However, the latent variables at different layers are distributed independently {\em a priori} and are only loosely connected. Instead, we propose a joint EBM prior that could tightly couple and jointly model latent variables at different layers and show superior results through joint training. 

\noindent\textbf{Hierarchical generator model with informative prior.} Recent advances \cite{cui2023learning,aneja2021contrastive} have started exploring learning informative prior for multi-layer generator models. JEBM \cite{cui2023learning} and NCP-VAE \cite{aneja2021contrastive} build latent space EBM on hierarchical generator models and showcase the capability of improving the generation quality. However, these models consider the \textit{conditional hierarchical models}, which can be limited in learning effective hierarchical representations (see Sec. \ref{section-bg-cond}). This work studies hierarchical representation learning by introducing a novel framework where latent space EBM can be jointly trained with the \textit{architectural hierarchical models} to capture complex representations of different levels.

\begin{figure*}[t]
\centering 
\includegraphics[width=0.95\textwidth]{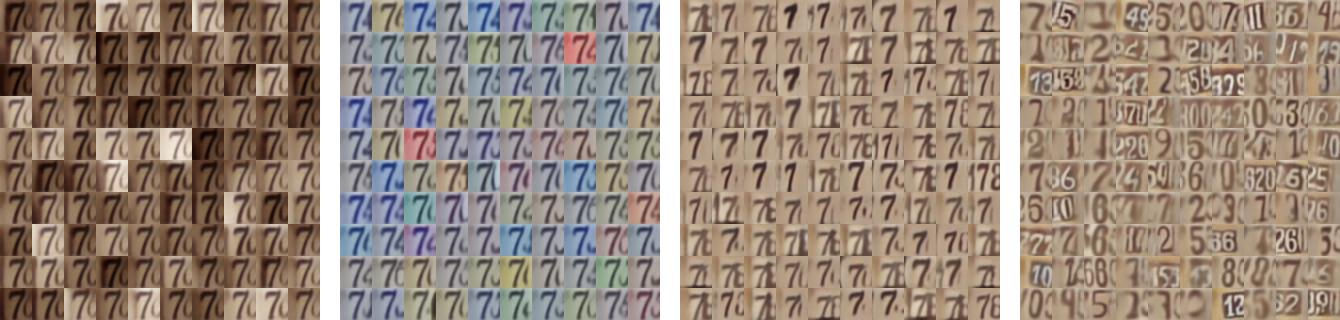} 
\caption{Hierarchical sampling on SVHN. \textbf{Left:} The latent code at bottom layer ($z_1$) represents the background light and shading. \textbf{Center-left:} the latent code at second bottom layer ($z_2$) represents the color schemes. \textbf{Center-right:} the latent code at second top layer ($z_3$) encodes the shape variations of the same digit. \textbf{Right:} the latent code at top layer ($z_4$) captures the digit identity and the general structure.} 
\label{Fig.svhn}
\end{figure*}

\begin{figure}[t]
\centering 
\includegraphics[width=0.90\columnwidth]{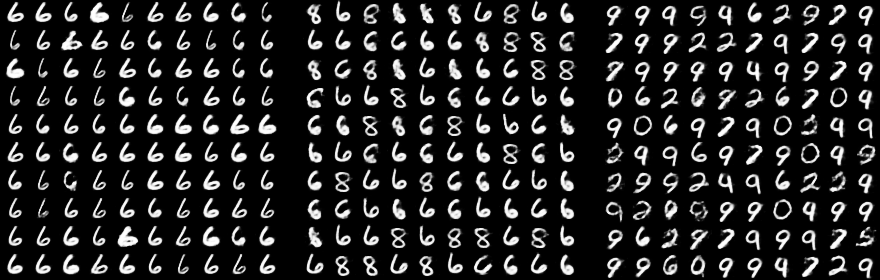} 
\caption{Hierarchical sampling on MNIST. \textbf{Left:} The latent code at bottom layer ($z_1$) indicates the stroke width.  \textbf{Center:} the latent code at second layer ($z_2$) encodes geometric changes among similar digits. \textbf{Right:} the latent code at top layer ($z_3$) learns the digit identity and general structure. } 
\label{Fig.mnist}
\end{figure}

\section{Experiments}
To demonstrate the effectiveness of our method, we conduct the following experiments: 1) hierarchical representation learning, 2) analysis of latent space, 3) image modelling, and 4) robustness. For a better understanding of the proposed method, we perform various ablation studies and an analysis of parameter complexity. We refer to additional experiments in Appendix.\ref{appendix-add}.

\subsection{Hierarchical Representation Learning}
\label{section-hierarchical-sampling}
We first demonstrate the capability of our model in hierarchical representation learning by experiments of hierarchical sampling and latent classifier.

\begin{figure}[h]
\centering 
\includegraphics[width=0.48\columnwidth]{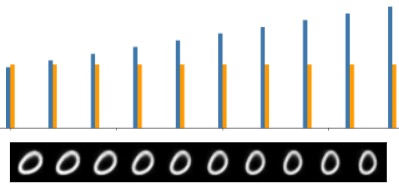}
\includegraphics[width=0.48\columnwidth]{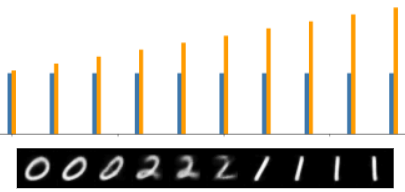}
\caption{Visualization for our model by changing each unit of 2-dimensional $\rvz$. \textbf{Left}: changing the first unit. \textbf{Right}: changing the second unit. \textbf{Top:} the value of each unit, where the orange color indicates the first unit and the blue color indicates the second unit.} 
\label{Fig.ours-mnist}
\end{figure}

\noindent\textbf{Hierarchical sampling.}  With partitioned latent variables and the hierarchical structure of generator, the latent variables at higher layers should capture semantic features, while the lower layers should capture low-level features. We train our model on MNIST using the same setting as in Sec. \ref{section-bg-lebm} with only `0', `1', and `2' digit classes available and the latent dimension set to 2. In Fig. \ref{Fig.ours-mnist}, by changing the $\rvz_1$ (at the bottom layer), our model can successfully deliver the variations in shape (low-level features), and changing the $\rvz_2$ (at the top layer) leads to the variations in digit class (high-level features). This indicates a successfully learned hierarchical representation.

Further, we train our model on MNIST with all digit classes and the more challenging SVHN dataset, and we generate the images by sampling (using Eq. \ref{langevin}) the latent code of one layer while keeping other layers fixed. The results shown in Fig. \ref{Fig.svhn} and Fig. \ref{Fig.mnist} also demonstrate the hierarchical representation learned by our model. 

\noindent\textbf{Latent Classifier.} We next show that our model can achieve better performance in hierarchical representation learning. Note that the inference model $q_\phi(\rvz|\rvx)$ should be learned to approximate the true posterior $p_\theta(\rvz|\rvx)$ of the generator, which integrates both the hierarchical inductive bias and the energy-based refinement. Therefore, to measure the learned hierarchical features, we learn classifiers on the inferred latent codes at different layers to predict the label of the data and measure the testing classification accuracy. The latent codes that carry rich high-level semantic features should achieve higher classification accuracy. 

\begin{table}[h]
\centering
\caption{Testing accuracy of inferred latent codes at each layer. We denote $(\downarrow)$ for the layer that has \textbf{unexpectedly lower} accuracy. }
\resizebox{0.95\columnwidth}{!}{
\begin{tabular}{lccccc}
\toprule
      &                       & $L=1$ & $L=2$ & $L=3$ & $L=4$\\
\midrule
\multirow{2}{*}{MNIST}          & Ours     & 32.60\%    & 58.32\%   & 67.64\%   & - \\
                                & VLAE     & 35.69\%    & 57.86\%  & 52.08\% $(\downarrow)$     & - \\
\midrule
\multirow{2}{*}{Fashion-MNIST}  & Ours     & 39.27\%     & 44.81\%  & 84.27\%    & - \\
                                & VLAE     & 52.43\%     & 35.65\% $(\downarrow)$    & 83.24\%    & - \\
\midrule
\multirow{2}{*}{SVHN}   & Ours &  21.00\%    & 25.29\%    & 30.58\%    & 86.63\% \\
                        & VLAE & 22.59\%    & 26.97\%    & 56.70\%     & 52.14\% $(\downarrow)$\\
\bottomrule
\end{tabular}
}
\label{table:clf_acc}
\end{table}

In practice, we first train our model on standard benchmarks, such as MNIST, Fashion-MNIST, and SVHN. Then, we train classifiers on the inferred latent codes at each layer. We consider VLAE \cite{zhao2017learning} as our baseline model, which learns a structural latent space with standard Gaussian prior. We show comparison results in Tab. \ref{table:clf_acc}, where our learned inference model $q_\phi(\rvz|\rvx)$ extracts varying levels of semantic features for different layers of latent codes. The top layer carries the most significant features, and the bottom layer carries the least. The baseline model that assumes Gaussian prior, however, extracted mixed semantic features across different layers. For fair comparisons, we use the same generation and inference structure as the baseline model\footnote{https://github.com/ermongroup/Variational-Ladder-Autoencoder}. The structure of the classifier contains two linear layers with a hidden dimension set to 256 and uses $\text{ReLU}$ activation function. The classifier has softmax as its output layer. We do not employ drop-out and normalization strategies.

\subsection{Image Modelling}
\label{section-generation-quality}
In this section, we evaluate our model in image modelling by measuring the quality of generated and reconstructed images. We consider the baseline models that assume Gaussian prior, such as Alternating Back-propagation (ABP) \cite{han2017alternating}, Ladder-VAE (LVAE) \cite{sonderby2016ladder}, BIVA \cite{maaloe2019biva}, Short-run MCMC (SRI) \cite{nijkamp2020learning}, and VLAE \cite{zhao2017learning}, as well as generator models with informative prior, such as RAE \cite{ghosh2019variational}, Two-stages VAE (2s-VAE) \cite{dai2019diagnosing}, NCP-VAE \cite{aneja2021contrastive}, Multi-NCP \cite{xiao2022adaptive} and LEBM \cite{pang2020learning}. To make fair comparisons, we follow the protocol in \cite{pang2020learning}.

\begin{table}[h]
\centering
\caption{Testing reconstruction by MSE, and generation evaluation by FID on SVHN and CelebA-64.}
\resizebox{0.95\columnwidth}{!}{
\begin{tabular}{l c c c c}
\toprule
 & \multicolumn{2}{c}{SVHN} & \multicolumn{2}{c}{CelebA-64}\\
Model  & MSE $(\downarrow)$ & FID $(\downarrow)$ & MSE $(\downarrow)$ & FID $(\downarrow)$\\
\hline
\hline
ABP & -     & 49.71 & -     & 51.50 \\
LVAE & 0.014 & 39.26 & 0.028 & 53.40\\
BIVA & 0.010 & 31.65 & 0.010 & 33.58\\
SRI  & 0.011 & 35.23 & 0.011 & 36.84\\
VLAE & 0.016 & 43.95 & 0.010 & 44.05\\
\midrule
2s-VAE & 0.019 & 42.81 & 0.021 & 44.40 \\
RAE & 0.014 & 40.02 & 0.018 & 40.95 \\
NCP-VAE & 0.020 & 33.23 & 0.021 & 42.07 \\
Multi-NCP & \textbf{0.004} & 26.19 & 0.009 & 35.38 \\ 
LEBM & 0.008 & 29.44 & 0.013 & 37.87\\
\midrule
Ours & 0.008 & \textbf{24.16} & \textbf{0.004} & \textbf{32.15}\\
\bottomrule
\end{tabular}
}
\label{table:1}
\end{table}

\noindent \textbf{Synthesis:} We use Fr$\acute{e}$chet Inception Distance (FID) \cite{NIPS2017_8a1d6947} to quantitatively evaluate the sample quality. In Tab. \ref{table:1}, it can be seen that our model obtains superior generation quality compared to the baseline models.

\noindent \textbf{Reconstructions:} To evaluate the accuracy of our inference model, we compute mean square error (MSE) \cite{pang2020learning,nijkamp2019learning} for testing reconstruction. As shown in Tab. \ref{table:1}, the proposed model leads to more accurate and faithful reconstructions. 

\subsection{Analysis of Latent Space}
\label{section-energy-landscape}
\begin{figure}[h]
\centering 
\includegraphics[width=0.90\columnwidth]{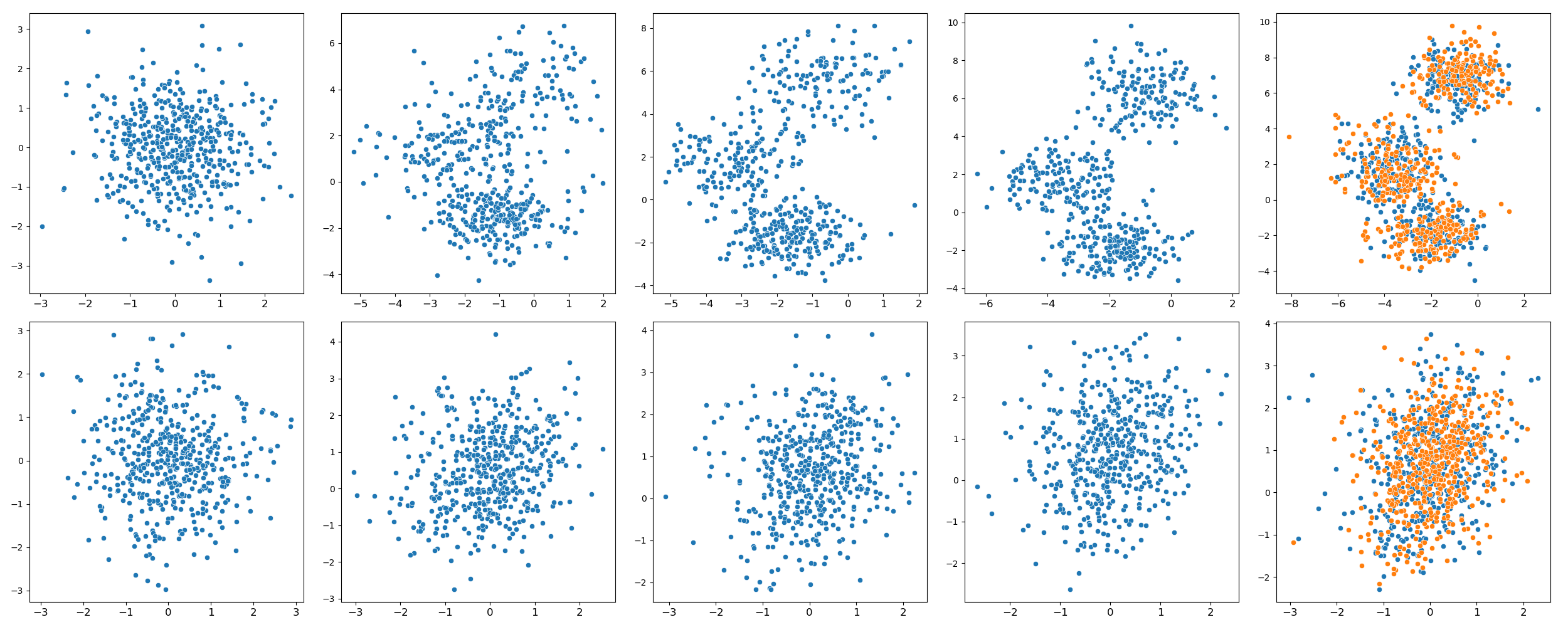} 
\caption{Visualization of the latent codes sampled from our EBM prior (Top row: $\rvz_2$). \textbf{Blue, Orange} color indicate prior and posterior, respectively. } 
\label{Fig.mnist-scatter-z}
\end{figure}

\noindent\textbf{Visualization of toy data.} To illustrate the expressivity of the proposed EBM prior, we train our model on MNIST (using the same setting as shown in Fig. \ref{Fig.lebm-mnist} and Fig. \ref{Fig.ours-mnist}) and set the latent dimension for both $\rvz_1$ and $\rvz_2$ to $2$ for better visualization. We visualized the Langevin transition of the prior sampling in Fig. \ref{Fig.mnist-scatter-z}, which shows that the latent codes at the top layer ($\rvz_2$) can effectively capture the multi-modal posterior, where using a Gaussian prior might be infeasible. 

\begin{figure}[h]
\centering
\includegraphics[width=0.99\columnwidth]{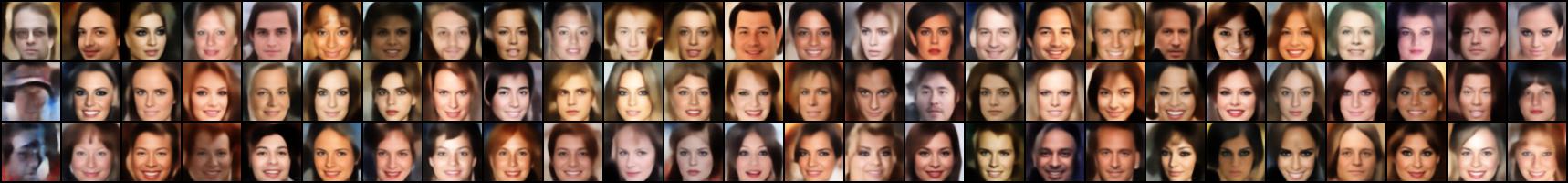}
\includegraphics[width=0.99\columnwidth]{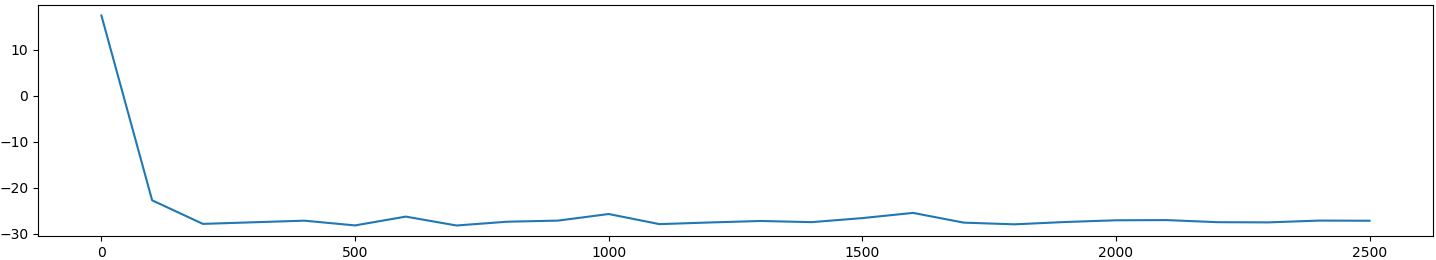}
\caption{Transition of Markov chains initialized from $p_0(\rvz)$ towards $p_{\alpha}(\rvz)$ for 2500 steps. Top: Trajectory in the CelebA-64 data space for every 100 steps. Bottom: Energy profile over time.}
\label{Fig.langevin}
\end{figure}

\noindent\textbf{Langevin trajectory.} We explore the energy landscape of the learned prior $p_{\alpha}(\rvz)$ via transition of Langevin dynamics initialized from $p_0(\rvz)$ towards $p_{\alpha}(\rvz)$ on CelebA-64. If the EBM is well-learned, the energy prior should render local modes of the energy function, and traversing these local modes should present diverse, realistic image synthesis and steady-state energy scores. Existing EBMs often suffer from oversaturated synthesis via the challenging \textit{long-run} Langevin dynamic (see oversaturated example in Fig.3 in \cite{nijkamp2020anatomy}). We thereby test our model with 2500 Langevin steps, which is much longer than the 40 steps used in training. Fig. \ref{Fig.langevin} shows the image synthesis and energy profile, where our EBM prior model does not exhibit the oversaturated phenomenon and successfully delivers diverse and realistic synthesis with steady-state energy.

\subsection{Robustness}
\label{section-adverarial-attack}
In this section, we examine the robustness of our model in adversarial attack and outlier detection.

\noindent\textbf{Adversarial robustness.} For adversarial robustness, we consider the latent attack \cite{gondim2018adversarial}, which is shown to be a strong attack for the generative model with latent variables, and it allows us to attack the latent codes at different layers. Specifically, let $\rvx_{t}$ be the target and $\rvx_{o}$ be the original image. The attack aims to create the distorted image $\tilde{\rvx}= \rvx_o + \epsilon$ such that its learned inference distribution $q_\phi(\rvz|\tilde{\rvx})$ is close to the inference $q_\phi(\rvz|\rvx_t)$ on target images under KL-divergence (Eq. 5 in \cite{gondim2018adversarial}).

\begin{figure}[h]
\centering
\includegraphics[width=0.22\columnwidth]{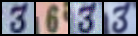}
\includegraphics[width=0.22\columnwidth]{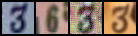}
\includegraphics[width=0.22\columnwidth]{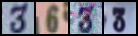}
\includegraphics[width=0.22\columnwidth]{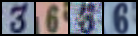}
\caption{Adversarial latent attack on different layers. The leftmost figure indicates the bottom layer, and the rightmost figure indicates the top layer. \textbf{1st col:} original image. \textbf{2nd col:} target image. \textbf{3rd col:} adversarial example. \textbf{4th col:} reconstruction of adversarial example.}
\label{Fig.latent_attack}
\end{figure}
\vspace{-0.5em}

We illustrate the adversarial attack on different layers in Fig. \ref{Fig.latent_attack}. The learned inference model $q_\phi(\rvz|\rvx)$ carries different levels of semantics for different layers. The adversarial example created by attacking latent codes at the bottom layer only perturbs the data along low-level features direction from the target image; hence, the model can be robust and have faithful reconstruction from adversarial examples. The adversarial example coming from the top-layer attack can distort the original data along a high-level semantic direction from the target image, thus, the reconstruction from the adversarial example can be substantially different from the original data, indicating a successful attack. 

\noindent\textbf{Anomaly detection.} We also evaluate the model on anomaly detection. If the model is well-learned, the inference model $q_\phi(\rvz|\rvx)$ should form an informative latent space that separates anomalies from the normal data. Following the protocols in \cite{kumar2019maximum,pang2020learning}, we take latent samples from the learned inference model and use the unnormalized log-posterior $\log p_{\theta}(\rvx, \rvz)$ as our decision function to detect anomaly on MNIST. Only one digit class is held-out as an anomaly in training, while both normal and anomalous data are used for testing. We compare baseline models such as BiGAN-$\sigma$ \cite{zenati2018efficient}, MEG \cite{kumar2019maximum}, LEBM \cite{pang2020learning}, HVAE and VAE using area under the precision-recall curve (AUPRC). The results of baseline models are provided by \cite{pang2020learning}, and we use $L^{>0}$ \cite{havtorn2021hierarchical} as the decision function for HVAE model. Tab. \ref{table.ad} shows the result.  

\begin{table}[h]
\caption{AUPRC scores for unsupervised anomaly detection on MNIST. Following the protocol of \cite{pang2020learning}, we average the results over last 10 epochs to account for variance.}
\resizebox{\columnwidth}{!}{%
\centering
\begin{tabular}{||c | c | c | c | c | c ||} 
 \hline
 hold-out digit & 1 & 4 & 5 & 7 & 9 \\ 
 \hline
 VAE & 0.063 & 0.337 & 0.325 & 0.148 & 0.104 \\ 
 HVAE & 0.494 $\pm$ 0.004  & 0.920 $\pm$ 0.004  & 0.913 $\pm$ 0.003  & 0.680 $\pm$ 0.006  & 0.791 $\pm$ 0.008 \\
 MEG & 0.281 $\pm$ 0.035  & 0.401 $\pm$ 0.061  & 0.402 $\pm$ 0.062  & 0.290 $\pm$ 0.040  & 0.342 $\pm$ 0.034 \\ 
 BiGAN-$\sigma$ & 0.287 $\pm$ 0.023  & 0.443 $\pm$ 0.029  & 0.514 $\pm$ 0.029  & 0.347 $\pm$ 0.017  & 0.307 $\pm$ 0.028 \\ 
 Latent EBM & 0.336 $\pm$ 0.008  & 0.630 $\pm$ 0.017  & 0.619 $\pm$ 0.013  & 0.463 $\pm$ 0.009  & 0.413 $\pm$ 0.010 \\
 Ours & \textbf{0.722 $\pm$ 0.010}  & \textbf{0.949 $\pm$ 0.002}  & \textbf{0.980 $\pm$ 0.001}  & \textbf{0.941 $\pm$ 0.003}  & \textbf{0.935 $\pm$ 0.003} \\
 \hline
\end{tabular}%
}
\label{table.ad}
\end{table}

\subsection{Ablation Studies}
\label{section-ablation-studies}
The proposed expressive prior model enables complex data representations to be effectively captured, which in turn improves the expressivity of the whole model in generating high-quality synthesis. To better understand the influence of the proposed method, we conduct ablation studies, including 1) MCMC steps, 2) complexity of EBM, and 3) other image datasets.

\noindent \textbf{MCMC steps:} We examine the influence of steps of short-run MCMC for prior sampling. Tab. \ref{table:5} shows that increasing steps could result in a better quality of generation. Increasing the step number from 15 to 60 could result in a significant improvement in synthesis quality, while exhibiting only minor influence when increased beyond 60.


\begin{table}[h]
\caption{Varying $k$ (steps number) of our model on CelebA-64.}
\centering
\resizebox{0.8\columnwidth}{!}{
\begin{tabular}{c c c c c c}
    \hline
                & $k=$15  & $k=$30 & $k=$60 & $k=$150 & $k=$300\\
    \hline
    FID         & 56.42   & 39.89  & \textbf{32.15}  & 31.20 & 30.78 \\
    \hline

\end{tabular}
}
\label{table:5}
\end{table}

\noindent \textbf{Complexity of EBM:} We further examine the influence of model parameters of EBM. The energy function is parameterized by a 2-layer perceptron. We fix the number of layers and increase the hidden units of EBM. As shown in Tab. \ref{table.ebm}, increasing the hidden units of EBM results in better performance.  

\begin{figure}[h]
\centering
\includegraphics[width=0.98\columnwidth]{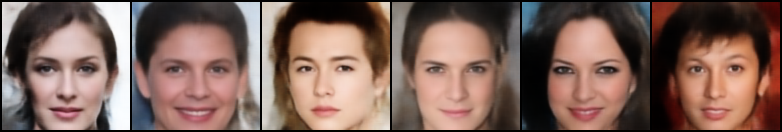}
\caption{Generated image synthesis on CelebA-128}
\label{Fig.Celeba128}
\end{figure}

\begin{table}[h]
\caption{Increasing hidden units (denoted as \textbf{nef}) of EBM.}
\centering
\resizebox{0.95\columnwidth}{!}{
\begin{tabular}{lccccc}
\hline
\textbf{nef}  & \textbf{nef}$=0$  & \textbf{nef}$=10$ & \textbf{nef}$=20$ & \textbf{nef}$=50$ & \textbf{nef}$=100$\\
\hline
FID     & 43.95 & 41.72  & 33.10   & 30.42  & \textbf{24.16}  \\
\hline
\end{tabular}
}

\label{table.ebm}
\end{table}
\vspace{-1em}
\noindent \textbf{Other image datasets:} We test the scalability of our model on high-resolution image dataset, such as CelebA-128. We show the image synthesis in Fig. \ref{Fig.Celeba128}. 
\vspace{-0.5em}

We also train our model on CIFAR-10 (32x32) and report the FID score for generation quality in Tab. \ref{tab:cifar10-fid}, in which our model performs well compared to baseline models. 
\begin{table}[h]
    \caption{FID on CIFAR-10}
    \centering
    \resizebox{0.8\columnwidth}{!}{
    \begin{tabular}{c c c c c c}
    \hline
    Ours & 2s-VAE & RAE & NCP-VAE & Multi-NCP & LEBM \\
    \hline
    63.42 & 72.90 & 74.16 & 78.06 & 65.01 & 70.15 \\
    \hline
    \end{tabular}
    }
    \label{tab:cifar10-fid}
\end{table}
\section{Conclusion}
In this paper, we propose to build a joint latent space EBM with hierarchical structures for better hierarchical representation learning. Our joint EBM prior is expressive in modelling the inter-layer relation among multiple layers of latent variables, which in turn enables different levels of data representations to be effectively captured. We develop an efficient variational learning scheme and conduct various experiments to demonstrate the capability of our model in hierarchical representation learning and image modelling.
For future work, we shall explore learning the hierarchical representations from other domains, such as videos and texts. 
{\small
\bibliographystyle{ieee_fullname}
\bibliography{egbib}

\begin{thebibliography}{10}\itemsep=-1pt

\bibitem{aneja2021contrastive}
Jyoti Aneja, Alex Schwing, Jan Kautz, and Arash Vahdat.
\newblock A contrastive learning approach for training variational autoencoder priors.
\newblock {\em Advances in neural information processing systems}, 34:480--493, 2021.

\bibitem{bachman2016architecture}
Philip Bachman.
\newblock An architecture for deep, hierarchical generative models.
\newblock {\em Advances in Neural Information Processing Systems}, 29, 2016.

\bibitem{3dshapes18}
Chris Burgess and Hyunjik Kim.
\newblock 3d shapes dataset.
\newblock https://github.com/deepmind/3dshapes-dataset/, 2018.

\bibitem{NEURIPS2018_1ee3dfcd}
Ricky T.~Q. Chen, Xuechen Li, Roger~B Grosse, and David~K Duvenaud.
\newblock Isolating sources of disentanglement in variational autoencoders.
\newblock In S. Bengio, H. Wallach, H. Larochelle, K. Grauman, N. Cesa-Bianchi, and R. Garnett, editors, {\em Advances in Neural Information Processing Systems}, volume~31. Curran Associates, Inc., 2018.

\bibitem{child2020very}
Rewon Child.
\newblock Very deep vaes generalize autoregressive models and can outperform them on images.
\newblock {\em arXiv preprint arXiv:2011.10650}, 2020.

\bibitem{cui2023learning}
Jiali Cui, Ying~Nian Wu, and Tian Han.
\newblock Learning joint latent space ebm prior model for multi-layer generator.
\newblock In {\em Proceedings of the IEEE/CVF Conference on Computer Vision and Pattern Recognition}, pages 3603--3612, 2023.

\bibitem{dai2019diagnosing}
Bin Dai and David Wipf.
\newblock Diagnosing and enhancing vae models.
\newblock {\em arXiv preprint arXiv:1903.05789}, 2019.

\bibitem{du2020improved}
Yilun Du, Shuang Li, Joshua Tenenbaum, and Igor Mordatch.
\newblock Improved contrastive divergence training of energy based models.
\newblock {\em arXiv preprint arXiv:2012.01316}, 2020.

\bibitem{du2019implicit}
Yilun Du and Igor Mordatch.
\newblock Implicit generation and generalization in energy-based models.
\newblock {\em arXiv preprint arXiv:1903.08689}, 2019.

\bibitem{gao2020learning}
Ruiqi Gao, Yang Song, Ben Poole, Ying~Nian Wu, and Diederik~P Kingma.
\newblock Learning energy-based models by diffusion recovery likelihood.
\newblock {\em arXiv preprint arXiv:2012.08125}, 2020.

\bibitem{ghosh2019variational}
Partha Ghosh, Mehdi~SM Sajjadi, Antonio Vergari, Michael Black, and Bernhard Sch{\"o}lkopf.
\newblock From variational to deterministic autoencoders.
\newblock {\em arXiv preprint arXiv:1903.12436}, 2019.

\bibitem{gondim2018adversarial}
George Gondim-Ribeiro, Pedro Tabacof, and Eduardo Valle.
\newblock Adversarial attacks on variational autoencoders.
\newblock {\em arXiv preprint arXiv:1806.04646}, 2018.

\bibitem{goodfellow2014generative}
Ian Goodfellow, Jean Pouget-Abadie, Mehdi Mirza, Bing Xu, David Warde-Farley, Sherjil Ozair, Aaron Courville, and Yoshua Bengio.
\newblock Generative adversarial nets.
\newblock {\em Advances in neural information processing systems}, 27, 2014.

\bibitem{gulrajani2016pixelvae}
Ishaan Gulrajani, Kundan Kumar, Faruk Ahmed, Adrien~Ali Taiga, Francesco Visin, David Vazquez, and Aaron Courville.
\newblock Pixelvae: A latent variable model for natural images.
\newblock {\em arXiv preprint arXiv:1611.05013}, 2016.

\bibitem{guo2018long}
Jiaxian Guo, Sidi Lu, Han Cai, Weinan Zhang, Yong Yu, and Jun Wang.
\newblock Long text generation via adversarial training with leaked information.
\newblock In {\em Proceedings of the AAAI Conference on Artificial Intelligence}, number~1, 2018.

\bibitem{han2017alternating}
Tian Han, Yang Lu, Song-Chun Zhu, and Ying~Nian Wu.
\newblock Alternating back-propagation for generator network.
\newblock In {\em Proceedings of the AAAI Conference on Artificial Intelligence}, volume~31, 2017.

\bibitem{havtorn2021hierarchical}
Jakob D~Drachmann Havtorn, Jes Frellsen, Soren Hauberg, and Lars Maal{\o}e.
\newblock Hierarchical vaes know what they don’t know.
\newblock In {\em International Conference on Machine Learning}, pages 4117--4128. PMLR, 2021.

\bibitem{NIPS2017_8a1d6947}
Martin Heusel, Hubert Ramsauer, Thomas Unterthiner, Bernhard Nessler, and Sepp Hochreiter.
\newblock Gans trained by a two time-scale update rule converge to a local nash equilibrium.
\newblock In I. Guyon, U.~V. Luxburg, S. Bengio, H. Wallach, R. Fergus, S. Vishwanathan, and R. Garnett, editors, {\em Advances in Neural Information Processing Systems}, volume~30. Curran Associates, Inc., 2017.

\bibitem{kingma2013auto}
Diederik~P Kingma and Max Welling.
\newblock Auto-encoding variational bayes.
\newblock {\em arXiv preprint arXiv:1312.6114}, 2013.

\bibitem{kumar2019maximum}
Rithesh Kumar, Sherjil Ozair, Anirudh Goyal, Aaron Courville, and Yoshua Bengio.
\newblock Maximum entropy generators for energy-based models.
\newblock {\em arXiv preprint arXiv:1901.08508}, 2019.

\bibitem{lecun2006tutorial}
Yann LeCun, Sumit Chopra, Raia Hadsell, M Ranzato, and F Huang.
\newblock A tutorial on energy-based learning.
\newblock {\em Predicting structured data}, 1(0), 2006.

\bibitem{li2020progressive}
Zhiyuan Li, Jaideep~Vitthal Murkute, Prashnna~Kumar Gyawali, and Linwei Wang.
\newblock Progressive learning and disentanglement of hierarchical representations.
\newblock {\em arXiv preprint arXiv:2002.10549}, 2020.

\bibitem{maaloe2019biva}
Lars Maal{\o}e, Marco Fraccaro, Valentin Li{\'e}vin, and Ole Winther.
\newblock Biva: A very deep hierarchy of latent variables for generative modeling.
\newblock {\em Advances in neural information processing systems}, 32, 2019.

\bibitem{neal2011mcmc}
Radford~M Neal et~al.
\newblock Mcmc using hamiltonian dynamics.
\newblock {\em Handbook of markov chain monte carlo}, 2(11):2, 2011.

\bibitem{nijkamp2020anatomy}
Erik Nijkamp, Mitch Hill, Tian Han, Song-Chun Zhu, and Ying~Nian Wu.
\newblock On the anatomy of mcmc-based maximum likelihood learning of energy-based models.
\newblock In {\em Proceedings of the AAAI Conference on Artificial Intelligence}, volume~34, pages 5272--5280, 2020.

\bibitem{nijkamp2019learning}
Erik Nijkamp, Mitch Hill, Song-Chun Zhu, and Ying~Nian Wu.
\newblock Learning non-convergent non-persistent short-run mcmc toward energy-based model.
\newblock {\em Advances in Neural Information Processing Systems}, 32, 2019.

\bibitem{nijkamp2020learning}
Erik Nijkamp, Bo Pang, Tian Han, Linqi Zhou, Song-Chun Zhu, and Ying~Nian Wu.
\newblock Learning multi-layer latent variable model via variational optimization of short run mcmc for approximate inference.
\newblock In {\em European Conference on Computer Vision}, pages 361--378. Springer, 2020.

\bibitem{pang2020learning}
Bo Pang, Tian Han, Erik Nijkamp, Song-Chun Zhu, and Ying~Nian Wu.
\newblock Learning latent space energy-based prior model.
\newblock {\em Advances in Neural Information Processing Systems}, 33:21994--22008, 2020.

\bibitem{rezende2014stochastic}
Danilo~Jimenez Rezende, Shakir Mohamed, and Daan Wierstra.
\newblock Stochastic backpropagation and approximate inference in deep generative models.
\newblock In {\em International conference on machine learning}, pages 1278--1286. PMLR, 2014.

\bibitem{Saito_2017_ICCV}
Masaki Saito, Eiichi Matsumoto, and Shunta Saito.
\newblock Temporal generative adversarial nets with singular value clipping.
\newblock In {\em Proceedings of the IEEE International Conference on Computer Vision (ICCV)}, Oct 2017.

\bibitem{NIPS2016_6ae07dcb}
Casper~Kaae S\o~nderby, Tapani Raiko, Lars Maal\o~e, S\o ren~Kaae S\o~nderby, and Ole Winther.
\newblock Ladder variational autoencoders.
\newblock In D. Lee, M. Sugiyama, U. Luxburg, I. Guyon, and R. Garnett, editors, {\em Advances in Neural Information Processing Systems}, volume~29. Curran Associates, Inc., 2016.

\bibitem{sonderby2016ladder}
Casper~Kaae S{\o}nderby, Tapani Raiko, Lars Maal{\o}e, S{\o}ren~Kaae S{\o}nderby, and Ole Winther.
\newblock Ladder variational autoencoders.
\newblock {\em Advances in neural information processing systems}, 29, 2016.

\bibitem{song2020score}
Yang Song, Jascha Sohl-Dickstein, Diederik~P Kingma, Abhishek Kumar, Stefano Ermon, and Ben Poole.
\newblock Score-based generative modeling through stochastic differential equations.
\newblock {\em arXiv preprint arXiv:2011.13456}, 2020.

\bibitem{takida2022sq}
Yuhta Takida, Takashi Shibuya, WeiHsiang Liao, Chieh-Hsin Lai, Junki Ohmura, Toshimitsu Uesaka, Naoki Murata, Shusuke Takahashi, Toshiyuki Kumakura, and Yuki Mitsufuji.
\newblock Sq-vae: Variational bayes on discrete representation with self-annealed stochastic quantization.
\newblock {\em arXiv preprint arXiv:2205.07547}, 2022.

\bibitem{tolstikhin2017wasserstein}
Ilya Tolstikhin, Olivier Bousquet, Sylvain Gelly, and Bernhard Schoelkopf.
\newblock Wasserstein auto-encoders.
\newblock {\em arXiv preprint arXiv:1711.01558}, 2017.

\bibitem{tomczak2018vae}
Jakub Tomczak and Max Welling.
\newblock Vae with a vampprior.
\newblock In {\em International Conference on Artificial Intelligence and Statistics}, pages 1214--1223. PMLR, 2018.

\bibitem{Tulyakov_2018_CVPR}
Sergey Tulyakov, Ming-Yu Liu, Xiaodong Yang, and Jan Kautz.
\newblock Mocogan: Decomposing motion and content for video generation.
\newblock In {\em Proceedings of the IEEE Conference on Computer Vision and Pattern Recognition (CVPR)}, June 2018.

\bibitem{vahdat2020nvae}
Arash Vahdat and Jan Kautz.
\newblock Nvae: A deep hierarchical variational autoencoder.
\newblock {\em Advances in Neural Information Processing Systems}, 33:19667--19679, 2020.

\bibitem{xiao2022adaptive}
Zhisheng Xiao and Tian Han.
\newblock Adaptive multi-stage density ratio estimation for learning latent space energy-based model.
\newblock In Alice~H. Oh, Alekh Agarwal, Danielle Belgrave, and Kyunghyun Cho, editors, {\em Advances in Neural Information Processing Systems}, 2022.

\bibitem{xiao2020vaebm}
Zhisheng Xiao, Karsten Kreis, Jan Kautz, and Arash Vahdat.
\newblock Vaebm: A symbiosis between variational autoencoders and energy-based models.
\newblock {\em arXiv preprint arXiv:2010.00654}, 2020.

\bibitem{xie2016theory}
Jianwen Xie, Yang Lu, Song-Chun Zhu, and Yingnian Wu.
\newblock A theory of generative convnet.
\newblock In {\em International Conference on Machine Learning}, pages 2635--2644. PMLR, 2016.

\bibitem{yu2017seqgan}
Lantao Yu, Weinan Zhang, Jun Wang, and Yong Yu.
\newblock Seqgan: Sequence generative adversarial nets with policy gradient.
\newblock In {\em Proceedings of the AAAI conference on artificial intelligence}, volume~31, 2017.

\bibitem{zenati2018efficient}
Houssam Zenati, Chuan~Sheng Foo, Bruno Lecouat, Gaurav Manek, and Vijay~Ramaseshan Chandrasekhar.
\newblock Efficient gan-based anomaly detection.
\newblock {\em arXiv preprint arXiv:1802.06222}, 2018.

\bibitem{zhao2017learning}
Shengjia Zhao, Jiaming Song, and Stefano Ermon.
\newblock Learning hierarchical features from generative models.
\newblock {\em arXiv preprint arXiv:1702.08396}, 2017.

\end{thebibliography}
}

\clearpage
\begin{alphasection}
\section{Additional Experiment}\label{appendix-add}
In order to better understand the proposed method, we conduct additional experiments, including (i) out-of-distribution detection and (ii) disentanglement learning, which further explore the potential of our method in various challenging tasks. 
\subsection{Out-of-Distribution Detection}
We conduct the out-of-distribution (OOD) detection for our model. We train our model on Fashion-MNIST (in-distribution) with MNIST (OOD) being the test dataset. We consider baseline models, including VLAE and HVAE, in which HVAE belongs to the \textit{conditional hierarchical model} (see Sec. 2.1), allowing a specialized decision function (likelihood-ratio, LLR \cite{havtorn2021hierarchical}) to be applied. It can be hard to directly apply such decision function to the our \textit{architectural hierarchical model}, thus we use the unnormalized log-posterior as the decision function (same as Anomaly Detection in Sec. 5.4) for our model and report the results in Tab. \ref{table.ood}. It can be seen that our model shows superior performance compared to VLAE, while HVAE could render better performance by using the specialized LLR decision function.

\begin{table}[h]
\small
\caption{Fashion-MNIST (in) vs MNIST (out)}
\setlength{\abovecaptionskip}{-0.cm}
\centering
\begin{tabular}{lccccc}
\hline
                          & $\text{HVAE(LLR$^{>k}$)}$  & $\text{HVAE}$ & $\text{VLAE}$ & $\textbf{Ours}$ \\
\hline
$\text{AUPRC}\uparrow$    & 0.984               & 0.363           & 0.344.           & 0.893\\
$\text{AUROC}\uparrow$    & 0.984               & 0.268           & 0.199            & 0.897          \\
\#param                   & 3.4M                & 3.4M            & 3.2M             & 3.2M\\
\hline
\end{tabular}
\label{table.ood}
\end{table}
\vspace{-1em}


\subsection{Disentanglement Learning}
\begin{figure}[h]
\centering
\includegraphics[width=0.84\columnwidth]{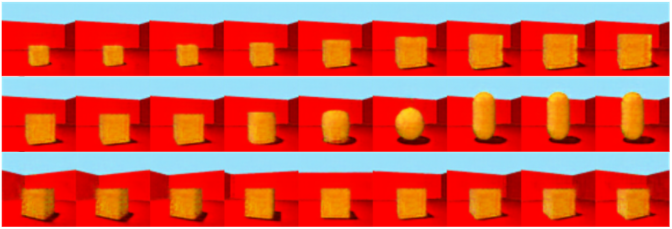}
\includegraphics[width=0.84\columnwidth]{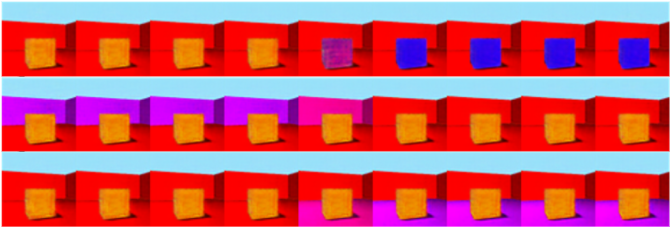}
\includegraphics[width=0.84\columnwidth]{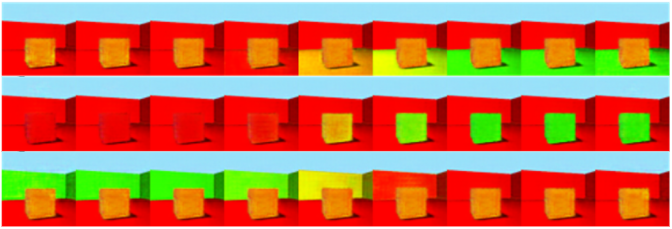}
\caption{Disentanglement traverse of each latent unit of our model. \textbf{Top panel (top 3 rows):} the top layer $\rvz_3$, and ${\rvz_3}_{dim}=3$. The top row of each panel illustrates the traverse on the first unit.}
\label{Fig.disentangle}
\end{figure}
We examine the disentanglement learning of our model. We train our model on 3DShapes dataset\cite{3dshapes18} with $L=3$ layers and visualize the traverse of each latent dimension in Figure \ref{Fig.disentangle}. We could see that the semantic factors, such as shape, size and direction, are disentangled into the latent dimensions of the top layer (i.e., $\rvz_3$), while low-level factors, such as background and object color, are disentangled into the latent dimensions of lower layers(i.e., $\rvz_1$, $\rvz_2$). 

\begin{table}[h]
\centering
\caption{MIG and MIG-sup on 3DShapes. $s$ denotes the progressive steps of pro-VLAE.}
\resizebox{0.98\columnwidth}{!}{
\begin{tabular}{|c|c|c|c|c|}
    \hline
                &  Ours           & pro-VLAE $s=3$ & pro-VLAE $s=2$ & pro-VLAE $s=1$\\
    \hline
    MIG         & \textbf{0.554}  & 0.357 & 0.339 & 0.247\\
    \hline
    MIG-sup     & \textbf{0.672}  & 0.406 & 0.333 & 0.136 \\
    \hline
\end{tabular}
}

\label{table.dis}
\end{table}

We further quantitatively evaluate our model. Prior work \cite{li2020progressive} applies the progressive learning strategy on VLAE to improve disentangled factor learning and computes MIG\cite{NEURIPS2018_1ee3dfcd} and MIG-sup\cite{li2020progressive} as the measurement. The pro-VLAE with $\beta=1$ (without progressive learning) is then considered as our baseline model. We train our model with the same inference and generator model and use the same latent dimension as the baseline model, and we compare with the pro-VLAE that uses multiple progressive steps (i.e., steps = 1, 2, 3). The numbers of pro-VLAE are obtained by the code\footnote{https://github.com/Zhiyuan1991/proVLAE}, and the comparison is shown in Table \ref{table.dis}.
\end{alphasection}
\end{document}